# Back to the Future for Dialogue Research: A Position Paper

Philip R. Cohen

Laboratory for Dialogue Research, Faculty of Information Technology, Monash University
philip.cohen@monash.edu

**Abstract**

This short position paper is intended to provide a critique of current approaches to dialogue, as well as a roadmap for collaborative dialogue research. It is unapologetically opinionated, but informed by 40 years of dialogue research. No attempt is made to be comprehensive. The paper will discuss current research into building so-called "chatbots", slot-filling dialogue systems, and plan-based dialogue systems. For further discussion of some of these issues, please see (Allen et al., in press).

**Chatbots**

The currently dominant approach to building dialogue systems, particularly those commonly referred to as "chatbots", is to train a neural network-based system "end-to-end" based on a large corpus of human-human dialogues, potentially supplemented with other information. Such chatbots have been developed by many researchers and corporations, either for the purpose of so-called "chit-chat" about arbitrary topics, or occasionally to engage in goal-directed transactional dialogues in limited domains (e.g., (Ghazvininejad et al., 2018)).

Without getting into the specifics of how these systems are constructed, it is apparent that such an approach provides a method for *imitating* a dialogue but not for *participating* in one. We can see the limitations of the approach from examples like these:

> **Human:** *what is your job ?* **Machine:** *i 'm a lawyer .*
> **Human:** *what do you do ?* **Machine:** *i 'm a doctor*
> (Vinyals & Le, 2015).
>
> **A**: I wish [...] would introduce another vegetarian option besides the shroomburger. It's delicious but kind of ridiculous.
> **B**: This is the best j.crew in the world. Try the lemonade.
> (Ghazvininejad et al., 2018).

In addition to being unable to track the topic of the conversation, a major missing aspect of human behavior is apparent from Google's suggested responses to emails, which have been trained over huge corpora of human-human conversations. Though the responses may be appropriate in many contexts, the user still needs to supply the *intention*. The primary characteristics of having an intention are that having an intention leads to action, and the agent is *committed* to achieving that action (Bratman, 1987; Cohen & Levesque, 1990). If the agent tries and fails, the agent will try again (if it believes the action is still possible). Neural net chatbots do not do this -- they do not attempt to achieve an action or effect with their utterances, and upon failure devise or plan another means to do so. I claim that to *participate* in a dialogue, one needs to have such intentions to achieve effects via communication. Intentions are intimately related to plans in that the agent plans to achieve its goals, and the elements of those plans become intentions. The plan-based approach to dialogue was begun 40 years ago (Bruce, 1975; Cohen & Perrault, 1979; Allen & Perrault, 1980; Perrault & Allen, 1980) and has seen occasional developments since then (e.g., the Artimis system (Sadek et al., 1997) and Allen et al's recent work (Galescu et al., 2018)). We will return to this approach below.

**Slot-filling Dialogue Systems**

Whereas chatbots are built to talk about any topic for which human-human conversational training data is available, a more domain-limited approach is to build so-called "task-oriented" dialogue systems. First, a bit of history. Task-oriented dialogue systems were originally intended to give *the user* instructions and guidance in performing an action, such as to assemble an object, such as an air compressor (Grosz, 1977). Current efforts have concentrated on a more limited objective, namely to get *the system* to perform some action for the user, such as to book a hotel or restaurant reservation. The core problem that is addressed is to fill out required and optional attribute-values (termed "slots") in an action schema or "frame" for example, the date, time, and number of people for a restaurant reservation. If an argument is missing, the system asks the user to supply it.

Dating back to the Gus system (Bobrow et al., 1977), slot-filling dialogue systems are a narrow slice of transactional dialogue systems. The influential POMDP-based reinforcement-learned slot-filling systems (e.g., (Young et al., 2013)) attempt to train an optimal dialogue policy using reinforcement learning in response to actions taken by a "user simulator" (Schatzman et al., 2007), perhaps bootstrapped via Wizard of Oz-derived data. That is, during training the dialogue system issues a response to the user simulator and is rewarded or penalized based on how the ultimate dialogue is evaluated. The Dialogue State Tracking Challenge (Williams et al., 2013) has created a number of slot-filling dialogue corpora, which have led many groups to build such systems.

Slot-filling dialogue systems are limited in a number of ways. First, the current approach to building these systems limits the set of logical forms the dialogue system can consider by avoiding complex meaning representations. For example, slot-filling systems can be trained to expect simple atomic responses like *"7pm"* to its question *"what time do you want to eat?"* However, the systems typically will not accept such reasonable but complex responses as *"not before 7pm,"* *"between 7 and 8 pm,"* or *"the earliest time available."* What's missing from these systems are true logical forms that employ a variety of operators (e.g., and, or, not, all, if-then-else, some, every, before, after, count, superlatives, comparatives, etc.). Many of these require scoped representations. For example, *"What is the closest parking to the Japanese restaurant nearest to the Space Needle?"* will have two superlative expressions, one embedded within the other. Furthermore, adequate meaning representations are compositional, such as the one underlying *"What are the three best Chinese or Japanese restaurants within walking distance of Century Link Field?"* They can also encode hypothetical information, as in *"If I take the American Airlines morning flight to Sydney, would I be able to arrive the next day?"* The evaluation of such expressions, with count, disjunction, and/or conditional operators, would prove problematic for neural network algorithms that merely "attend" to a knowledge base rather than evaluate the logical form. Among the other difficulties encountered by such systems are so-called "multi-intent" utterances that essentially require database joins. However, complex logical forms such as these can now be produced robustly from competent neural network semantic parsers (e.g., Duong et al., 2017; Wang et al., 2015).

Second, reinforcement-learned dialogue systems are limited by the capabilities of the user simulators used to train them. Typically, these simulators have a stack of goals that are derived from the operations of slot-filling, either to inform the user of the result of an action, or to request the user to provide a value for a slot. However, there are many more types of goals involved in dialogues (Galescu et al., 2018) such as to abandon a goal, re-attempt a goal, etc. Furthermore, many utterances one might reasonably ask do not involve specific slots per se, but rather involve constraints that the slot values need to satisfy. The current simulators do not support generating such utterances, and thus because they do not provide the requisite training data, the dialogue system cannot learn the behavior.

Third, the dialogue initiative is rigidly prescribed. Essentially, the user makes a request, the system asks for the missing information, the user supplies that information, the system (eventually) confirms the action to be performed, the user agrees or disconfirms, etc. However, real dialogues can have many shifts of initiative in which the parties collaborate to accomplish goals.

To put these missing capabilities in one example that shows the need for a more powerful logical form language and for mixed initiative dialogue processing, consider this conversation:

1. U: Please book a reservation at the closest good restaurant to the Orpheum Theater on Monday for 4 people.
2. S: OK, what time would you like to eat?
3. U: what's the earliest time available?
4. S: 6 pm
5. U: too early
6. S: how about 7 pm?
7. U: OK
8. S: OK, shall I make a reservation at Guillaume for 4 people at 7pm?
9. U: Sure.
10. S: Done!
11. U: Can you let Sharon know?
12. S: What is her phone number?
13. U: It's 650-123-4567
14. S: OK, sending text to Sharon that you have a reservation at Guillaume on Monday for 4 people.

Here the system has responded to the user's complex request with a slot-filling question (2), the effect of which is that the user believes the system wants to know the time that the user wants to eat. Rather than answer the question, the user replies with another question (3), a not infrequent occurrence though it violates the typical assumptions of simple dialogue systems. Notice that Question (3) starts a subdialogue (3-7) to find the time the user wants to eat. Question (3) is intended to establish a *constraint* on the desired time in the effect of (2). The times specified by the system in (4) and (6) are not times the user wants to eat but proposed *available* times, the latter of which is confirmed as the *desired time* by the user in (7). The disconfirm speech act in (5) indicates that the time proposed by the system is not desired. However, the slot-filling approach assumes that it is *the user* who fills the slots that have been requested, rather than confirming values that have proposed by the system. In fact, any subdialogue that results in the state in which the system knows the time the user wants to eat should be acceptable.

Utterance (8) is multifunctional in that it informs the user of the restaurant that the system has found to satisfy the user's constraints, and requests permission to book it. What matters here are the goals being accomplished, not the dialogue act labels themselves. The indirect request in (11) assumes a propositional anaphor (the booking event) as the content of the requested informative action. Then in (12) the system asks a slot-filling question whose answer it needs to know in order to perform an inferred action that satisfies the indirect request in (11). Unlike exchange (2)-(3), this time a literal question (which is interpreted as a request) from *the user* is followed by a question from the system, which shows again a shift of initiative. Current approaches have yet to be able to handle reasonable dialogues such as this. We argue that an approach based on plans/goals/intentions that reason over the effects of speech acts operating on proper logical forms can derive the appropriate communicative acts in such dialogues.

**Plan-based Model of Dialogue** Dialogue researchers have long emphasized that conversation is best analyzed as a special case of plan-based collaborative behavior (Grosz & Sidner, 1990). Plan-based interaction acknowledges that communication is a special case of purposeful behavior (Allen & Perrault, 1980; Cohen & Perrault, 1979). Collaborative interaction involves agents being jointly committed to their partners' success (Cohen & Levesque, 1991). In doing so, an agent recognizes its partner's plans to achieve the joint goal, and then performs actions to facilitate them. People have learned to be helpful at a very young age and are strongly expected to collaborate as part of ordinary social interaction (Warneken & Tomasello, 2006; see video). In general, people's plans involve physical (and now digital) acts, as well as speech acts (such as requests, questions, confirmations, etc.) When the process of collaboration is applied to communication, people infer the reasons behind their interlocutor's utterances and attempt to ensure their success by (at least) telling them what they need to know to be successful, and by potentially volunteering to perform actions on their behalf. Such reasoning is apparent when a system responds to the user's asking "Where is Dunkirk playing?" with "It's playing at the Roxy theater *at 7:30pm, however it is sold out*." Here the literal and truthful answer (shown here in plain font) is not sufficient and a respondent who knew that the theater was sold out would be considered uncooperative. A collaborator, on the other hand, will go beyond inferring the user's plan by attempting to debug that plan. If the plan is expected to fail, the collaborator may develop and suggest (or execute) an alternative plan to achieve the user's higher-level goal. To continue the movie example, a collaborative assistant system might then say *"It's also showing at the Forum theater tomorrow at 8pm, and tickets are available. Would you like me to purchase them?"* In order to provide such responses, the system needs to infer that people may want to know where an entity is (the location where the movie is showing), in order to go there (the theater), in order to perform a normal activity done on that entity at that location (watch a movie). The system checks the plan's preconditions (watching a movie requires that the agent has a ticket), and also the applicability conditions (tickets must be available). If the latter fails, the intention is impossible so the system will drop it and attempt to find another plan to achieve the higher level goal (of having seen the movie). The prototype system we would demonstrate at the workshop engages in the reasoning above. Such helpful conversational behavior is a paradigmatic example of *human-AI collaboration.*

However, except for hand-built examples, current virtual assistant systems are not typically providing such assistance. In order to build systems that can engage humans in collaborative plan-based dialogue, research is needed on planning, plan recognition, and reasoning about people's mental and social states (beliefs, desires, goals, intentions, permissions, obligations, etc.), and their relation to conventional behavior. Plan recognition involves observing actions and inferring the (structure of) reasons why those actions were performed, often to enable the actor to perform still other actions (Allen & Perrault, 1980; Geib & Goldman, 2009; Sukthankar, et al., 2014). Belief-desire-intention (BDI) theory (Cohen & Levesque, 1990) and architectures (Bratman et al., 1988) within the subfield of Multi-Agent Systems are intimately related to dialogue processing. Prior research, including our own, has developed prototypes of the above collaborative processing, and has shown that such collaborative BDI architectures and epistemic reasoning can form the basis for dialogue managers (Allen & Perrault, 1980; Cohen & Perrault, 1979; Sadek et al., 1997).

However, though the BDI approach has been researched for many years, with few exceptions (e.g., Allen's group at the University of Rochester/IHMC (Galescu et al., 2018)), it has not seen recent system application to collaborative dialogue. Our recent plan-based dialogue manager prototype (which I will demonstrate at the workshop) uses the same planner to reason about physical and speech acts, enabling the system to plan yes/no and wh-questions when the user is believed to know the answers[1], to make requests when the system wants the effect and the user is believed to be able to perform the requested action, to inform the user of facts the user is not believed to already know, to suggest actions that the user may want that would further the user's goals, etc.

**Concluding Remarks**

Our overall multi-year program of research is to build a scalable collaborative dialogue component based on planning/plan recognition. In order to build a collaborative architecture for use with communicative actions and that reasons about users' mental states (beliefs, goals, intentions), we are investigating probabilistic planning and plan recognition algorithms (Geib & Goldman, 2009) along with epistemic, intentional, and constraint reasoning.

Because the dialogue manager operates at the level of plans and goals as applied to physical, digital, and communicative acts, it can be domain independent. The logical forms encode domain dependent predicates and actions, as derived from the backend APIs, database schemas, and knowledge base (Duong et al., 2017; Wang et al., 2015).

---

[1] Note that the system needs to represent *that* the user knows the answer without representing *what* the user thinks the answer to be (or it would not need to ask). Such reasoning would be important in deciding whom to ask a question. Such "quantifying-in" (Allen & Perrault, 1980; Cohen & Perrault, 1979; Kaplan, 1968; Quine, 1956) rules out using simple databases to represent a user's belief state.

The domain actions that the planner/plan recognizer operate over can be discovered by crowd-sourcing and text mining (Fast et al., 2016; Jiang & Riloff, 2018).

The plan/plan recognition component could either be the engine of a complete dialogue manager, or it could generate possible response plans that then could be used to train a dialogue management component, similar to the "dialogue self-play" approach of (Shah et al., 2018). Thus, simulated dialogues that are situationally relevant could be created, and once paraphrased by a crowd with actual utterances, could be used for training (Duong et al., 2017; Shah et al., 2018; Wang et al., 2015). A plan-based dialogue manager that performs both planning and plan recognition could thus play both sides of a collaborative conversation by planning and interpreting speech acts and their propositional content, given parties' differing beliefs about the world state, goals and intentions. In this way, a system could learn how to reason, plan, and converse.

Finally, we are interested in systems that can *explain* their actions. Such systems should be able to answer questions such as "why did you say that?" Because a plan-based dialogue system has created plans that stand behind its utterances, it can explain what the utterance(s) were intended to achieve.

## References

Allen, J. F. and Perrault, C. R., Analyzing intention in utterances, *Artificial intelligence 15 (3),* 143-178.

Allen, J. F., André, E., Cohen, P. R., Hakkani-Tür, D., Kaplan, R., Lemon, O., Traum, D., Challenge discussion: Advancing multimodal dialogue, in *Handbook of Multimodal-Multisensor Interfaces*, Oviatt, S. L., Schuller, B., Cohen, P. R., Sonntag, D., Potamianos, G., and Krüger, A., ACM Press/Morgan & Claypool Publishers, in press.

Bobrow, D. G., Kaplan, R. M., Kay, M., Norman, D. A., Thompson, H., and Winograd, T. GUS, a frame-driven dialog system. *Artificial Intelligence, 8(2),* 1977, 155-173.

Bratman, M. E., Israel, D. and Pollack, M. E., Plans and resource-bounded practical reasoning. *Comp. Intelligence*, *4*:349-355, 1988.

Bratman. M. E., *Intentions, Plans, and Practical Reason*, Harvard University Press, Cambridge, MA, 1987.

Bruce, B. C., Generation as a social action, *Proc. of the Workshop on Theoretical Issues in Natural Language Processing*, ACL, 1975.

Cohen, P. R. and Levesque, H. J., Intention is choice with commitment, *Artif. Intelligence, 42 (2-3),* 1990, 213-261.

Cohen P. R. and Levesque, H. J. Teamwork, Noûs, 1991.

Cohen, P. R. and Perrault, C. R., Elements of a plan-based theory of speech acts, *Cognitive Science, 3(3),* 1979.

Duong, L., Afshar, H., Estival, D., Pink, G., Cohen, P. R., and Johnson M. Multilingual Semantic Parsing and Code-switching, *Proc. of the 21st Conf. on Computational Natural Language Learning (CoNLL 2017)*, 2017, pp. 379-389.

Fast, E., McGrath, W., Rajpurkar, P. and Bernstein, M. Augur: Mining Human Behaviors from Fiction to Power Interactive Systems. *Proc. of the 2016 CHI Conference on Human Factors in Computing Systems*, ACM Press, 2016.

Galescu, L., Teng, C. M., Allen J. F., and Pereira, I. Cogent: A Generic Dialogue System Shell Based on a Collaborative Problem Solving Model, *Proceedings of SigDial,* Assoc. for Computational Linguistics, 2018, 400-409.

Geib, C., and Goldman, R.P., A probabilistic plan recognition algorithm based on plan tree grammars, *Artificial Intelligence,* 2009.

Ghazvininejad, M., Brockett, C.,Chang, M-W., Dolan, B., Gao, J.,Yih, W-T, Galley, M. A Knowledge-Grounded Neural Conversation Model, *Proc. of AAAI,* 2018.

Grosz, B. J., The structure of task-oriented dialogues, *Proc. of IEEE Speech Symposium*, Carnegie Mellon University, 1974.

Grosz, B. J., & Sidner, C., Plans for discourse, in *Intentions in Communication*, Cohen, P. R., Morgan, J., and Pollack, M. E., MIT Press, 1990.

Jiang, T., and Riloff, E., Learning prototypical goal activities for locations, *Proc. of Assoc. for Comp. Ling.,* 2018, 1297-1307.

Kaplan, D. Quantifying in, *Synthese 19(1/2)*, 1968, 178-214.

Perrault, C. R. and Allen, J. F., A plan-based analysis of indirect speech acts, *Comp. Ling., 6(3-4),* 1980, 167-182.

Quine, W. V. O. Quantifiers and propositional attitudes, *Journal of Philosophy 53(5)*, 1956, 177-187.

Sadek, D., Bretier, P., and Panaget, F., ARTIMIS: Natural dialogue meets rational agency, *Proc. IJCAI-15*, 1997, pp. 1030-1035.

Shah, P., Hakkani-Tür, D., Tür, G., Rastogi, A., Bapna, A., Nayak, N., Heck, L., Building a conversational agent overnight with dialogue self-play, *arXiv: 1801.04871v1*, Jan., 2018.

Sukthankar, G., Geib, C., Bui, H., Pynadath, D., and Goldman, R., *Plan, Activity, and Intent Recognition: Theory and Practice*, San Francisco: Morgan Kauffman Publishers, 2014.

Vinyals, O., and Le, Q. A neural conversational model. *Proc. of ICML*, 2015.

Wang, Y., Berant, J., and Liang, P., Building a semantic parser overnight, *Proc. of Assoc. for Comp. Ling.*, 2015, 1332–1342.

Warneke, F. and Tomasello, M. Altruistic Helping in Human Infants and Young Chimpanzees, *Science 311(5765)* 03 Mar 2006, 1301-1303 (2006). https://www.youtube.com/watch?v=Z-eU5xZW7cU

Williams, J., Raux, A., Ramachandran, D. and Black A.The Dialog State Tracking Challenge, Proc. of SIGDIAL 2013, Assoc. for Comp Linguistics, 2013, 404-413.

Young, S., Gasic, M., Thomson, B., and Williams, J. D. POMDP-based Statistical Spoken Dialogue Systems: A review. *Proc. IEEE, 101(5),* 2013, 1160-1179.